%% file: root.tex
\documentclass[conference,10pt]{IEEEtran}
 \IEEEoverridecommandlockouts
\usepackage{cite}
\usepackage{color}
\ifCLASSINFOpdf
	\usepackage[pdftex]{graphicx}
	\graphicspath{{./figures/}}
\else
	\usepackage[dvips]{graphicx}
	\graphicspath{./figures/}
\fi
\graphicspath{{fig/}{jpeg/}}
\usepackage[cmex10]{amsmath}
\usepackage{amssymb}
\usepackage{algorithm}
\usepackage{algorithmic}
\usepackage{subcaption}
\usepackage{caption}
\usepackage[outdir=./]{epstopdf}
\usepackage{comment}
\usepackage{bm}
\input{mysymbol.sty}
\input{my_sections.tex}

\usepackage{tikz}
\usetikzlibrary{shapes,arrows}
\usepackage{float,environ}
\usetikzlibrary{matrix} 
\usetikzlibrary{decorations.markings} 
\usetikzlibrary{calc} 
\usetikzlibrary{decorations.pathmorphing,snakes, positioning, decorations.pathreplacing}

\usetikzlibrary{fit}

\usepackage{pgfplots}
\usepackage{color}
\usepackage{multirow}

\makeatletter
\newsavebox{\measure@tikzpicture}
\NewEnviron{scaletikzpicturetowidth}[1]{%
  \def\tikz@width{#1}%
  \begin{lrbox}{\measure@tikzpicture}%
  \BODY
  \end{lrbox}%
  \pgfmathparse{#1/\wd\measure@tikzpicture}%
  \BODY
}
\makeatother

\newcommand{\QED}{\hfill\ensuremath{\blacksquare}}

\def\vec2{\text{vec}}

\newtheorem{remark}{\hspace{0pt}\bf Remark}

\begin{document}
%
\title{Stochastic Sequential Decision Making \\ over Expanding Networks with Graph Filtering
}

\author{Zhan Gao$^{ \dagger}$, Bishwadeep Das$^{\ddagger }$ and Elvin Isufi$^{\ddagger }$
\thanks{$^\dagger$Department of Computer Science and Technology, University of Cambridge, Cambridge, UK. $^{\ddagger}$Faculty of EEMCS, Delft University of Technology, 
The Netherlands. Part of this work was supported by the TU Delft AI Labs programme, the NWO OTP GraSPA proposal \#19497, and the NWO VENI proposal 222.032. E-mails: zg292@cam.ac.uk, $\{$e.isufi-1, b.das$\}$@tudelft.nl }}

\maketitle

\begin{abstract}
Graph filters leverage topological information to process networked data with existing methods mainly studying fixed graphs, ignoring that graphs often expand as nodes continually attach with an unknown pattern. The latter requires developing filter-based decision-making paradigms that take evolution and uncertainty into account. Existing approaches rely on either pre-designed filters or online learning, limited to a myopic view considering only past or present information. To account for future impacts, we propose a stochastic sequential decision-making framework for filtering networked data with a policy that adapts filtering to expanding graphs. By representing filter shifts as agents, we model the filter as a multi-agent system and train the policy following multi-agent reinforcement learning. This accounts for long-term rewards and captures expansion dynamics through sequential decision-making. Moreover, we develop a context-aware graph neural network to parameterize the policy, which tunes filter parameters based on information of both the graph and agents. Experiments on synthetic and real datasets from cold-start recommendation to COVID prediction highlight the benefits of using a sequential decision-making perspective over batch and online filtering alternatives. 
\end{abstract}


\section{Introduction}

Graph filters are fundamental for processing networked data \cite{sandryhaila2013discrete, berberidis2018adaptive, isufi2024graph}, with widespread deployment and success in communications \cite{saad2020accurate, gao2025graph}, recommender systems \cite{shen2021powerful, peng2024powerful}, and robot control \cite{gao2022wide, gao2023online}. By modeling networked data as graph signals, filtering collects neighborhood information to extract features, and forms the backbone of more powerful solutions such as graph filter banks, graph wavelets, and graph neural networks (GNNs) \cite{gama2019convolutional,gama2020graphs}. The majority of graph filtering literature considers static graphs with fixed nodes and edges \cite{dong2020graph,dong2019learning}. However, real-world systems often contain dynamic graphs \cite{isufi2017filtering, gao2021stochastic, money2023sparse}, which pose challenges and demand adaptability of graph filters to the dynamics. This work focuses on continually expanding graphs with a growing number of nodes \cite{erdos_evolution_1961}, where new nodes attach to the existing graph with a fair degree of uncertainty, such as collaborative filtering \cite{liu2020heterogeneous} and growth in social, technological, and biological networks \cite{barabasi2016network}.

Filtering over expanding graphs is challenging as the number of nodes grows and the topology changes continuously. Re-training every time incurs significant computational overhead, rendering it impractical for real-time inference. The works in \cite{das2022learning, dasfiltering2020} approached this setting by building a stochastic attachment model adapted to performing interpolation or learning a filter, but they are limited to a single incoming node rather than a stream of nodes. In contrast, \cite{das2024online} designed online graph filters in the streaming setting and \cite{liu2018streaming} studied streaming graphs to generate an embedding based on a constrained optimization model. The works in \cite{chen2014semi, shen2019online, zong2021online} learn over expanding graphs by leveraging the connectivity of the incoming nodes and their features.  
For example, \cite{chen2014semi} classified an incoming node by using its feature information. 
The work in \cite{shen2019online} performed online node regression by constructing random kernel features based on the connectivity pattern, 
while \cite{zong2021online} extended to consider multi-hop information. However, this incoming node information may be unavailable or non-deterministic. Within the realm of GNNs, \cite{rongdropedge, gao2021training, schlichtkrullinterpreting, chen2025adedgedrop} studied dynamic graphs to avoid overfitting and save computation, but the graph remains the same size. The work in \cite{zhang2024topology} considered expanding graphs but focused on memory reduction, while \cite{cervino2023learning} introduced a learning technique that trains GNNs on growing graphs under the graphon family assumption. 

However, these works view the problem from a myopic perspective, relying solely on present or past information while overlooking the potential future impact during graph expansion. In this context, filter parameters are updated online with domain uncertainty, ultimately, propagating this uncertainty in the long-range decision chain. This motivates to study filtering over expanding graphs from a non-myopic perspective of stochastic sequential decision making. The latter enables to capture expansion dynamics and interpret information propagation, where actions and predictions at a particular instant affect future inference on successively expanding graphs. 

More specifically, we develop a stochastic sequential decision making framework over expanding graphs. Such a framework accounts for future impacts through long-term rewards across time steps, captures inherent dynamics of graph expansion, and adapts the filter to expanding graphs following reinforcement learning. Our detailed contributions are: 

\smallskip
\noindent \textbf{(i)} We formulate a stochastic sequential decision making problem over expanding graphs. It leverages a graph filter as the fundamental model that processes networked data to perform inference and a control policy that tunes the filter parameters under sequential stochastic constraints (Sec. \ref{sec_problem}). 

\smallskip
\noindent \textbf{(ii)} We propose to learn the policy following multi-agent reinforcement learning (MARL). By modeling filter shifts as agents and their parameters as states, we represent the filter as a multi-agent system. Each agent deploys a policy that generates an action to update its state, adapting the filter to the expanded graph. This captures graph expansion dynamics via a Markov decision process and incorporates future impacts in long-term rewards of MARL (Sec. \ref{subsec:MARL}). 

\smallskip
\noindent \textbf{(iii)} We develop a Context-aware Graph Neural Network (C-GNN) to parameterize the policy for MARL. C-GNN extracts context features from expanded graphs, concatenates context features and agent states, and leverages the latter to compute actions, thereby making decisions based on information of both the graph and agents (Sec. \ref{subsec:CGNN}). 

Experiments on synthetic and real data from recommender systems to COVID prediction validate our framework, and demonstrate improved performance and robust generalization.


\section{Problem Formulation}\label{sec_problem}

In this section, we formalize the setting of filtering over expanding graphs and formulate the problem of stochastic sequential decision making within this setting. 

\subsection{Filtering over Expanding Graphs}\label{subsec:graph}

Consider an initial graph $\ccalG_0$ with a node set $\ccalV_0=\{v_i\}_{i=1}^{N_0}$ and an edge set $\ccalE_0$. In the expanding setting, a sequence of $T$ incoming nodes $\{v_{N_t}\}_{t=1}^T$ arrive at times $t=1,...,T$, respectively. Node $v_{N_t}$ connects to the existing graph $\ccalG_{t-1}$, forms the expanded graph $\ccalG_t$ with $N_t = N_{t-1} + 1$, and stays attached. The initial topology is represented by a weighted adjacency matrix $\bbA_0 \in \mathbb{R}^{N_ 0 \times N_0}$. The topology $\bbA_t$ at time $t$ takes the form of
\begin{align}\label{eq:shiftOperator}
	\bbA_t = \begin{bmatrix}
		\bbA_{t-1} & \bba_{t-1} \\
		\bba_{t-1}^\top & 0
	\end{bmatrix},
\end{align}
where $\bba_{t-1} = [a_1,...,a_{N_{t-1}}]^\top$ is the attachment vector of the incoming node $v_{N_{t}}$. Each non-zero element of $\bba_{t-1}$ implies an edge between $v_{N_{t}}$ and an existing node in $\ccalV_{t-1}$. We focus on undirected edges representing bidirectional influence. In many real-world settings, $\bba_{t-1}$ is captured via a stochastic model \cite{barabasi1999emergence,liu2018streaming}, where $v_{N_{t}}$ attaches to $v_i\in\ccalV_{t-1}$ with a probability $p_i$. This lends itself nicely to growing models in network science \cite{barabasi2016network, barabasi1999emergence}, where attaching probabilities depend on the structure of $\bbA_{t-1}$. Under this setting, a graph signal processing task is to infer the signal value of the incoming node $x_{N_{t}}$ before it is observed, based on the signal values of the existing nodes $\bbx_{t-1} \in \mathbb{R}^{N_{t-1}}$ and their topological connections with the incoming node $\{\bbA_{t-1}, \bba_{t-1}\}$. 

Graph filters are suitable tools in processing signals for inference \cite{isufi2024graph}. Adapted to our setup, we define a zero-padded signal $\tilde{\bbx}_t = [\bbx_{t-1}, 0]^\top$, where $0$ indicates that $x_{N_t}$ is unknown. The filter of order $K$ takes $\tilde{\bbx}_t$ as input and outputs $\tilde{\bby}_t$ as
\begin{align}\label{eq:filter}
	\tilde{\bby}_t = \sum_{k=0}^K h_{t,k} \bbA_t^k \tilde{\bbx}_t := \ccalF(\bbA_t, \tilde{\bbx}_t, \bbh_t),
\end{align}
where $\bbh_t = [h_{t,0},...,h_{t,K}]^\top$ are filter parameters. By substituting \eqref{eq:shiftOperator} and $\tilde{\bbx}_t$ into \eqref{eq:filter}, we use the $N_t$th entry of the output 
\begin{align}\label{eq:prediction}
	[\tilde{\bby}_t]_{N_t} = \bba_{t-1}^\top \sum_{k=1}^K h_{t,k} \bbA_{t-1}^{k-1} \bbx_{t-1}
\end{align}
to predict $x_{N_t}$, and measure the prediction performance with a loss function $l([\tilde{\bby}_t]_{N_t}, x_{N_{t}})$. As the graph expands, the filter needs to be updated continuously. Online learning is a potential solution, but it updates the filter parameters on the fly using solely current or past information \cite{das2024online}. 

\begin{figure*}%
	\centering
	\begin{subfigure}{0.85\columnwidth}
		\includegraphics[width=1.\linewidth, height = 0.45\linewidth]{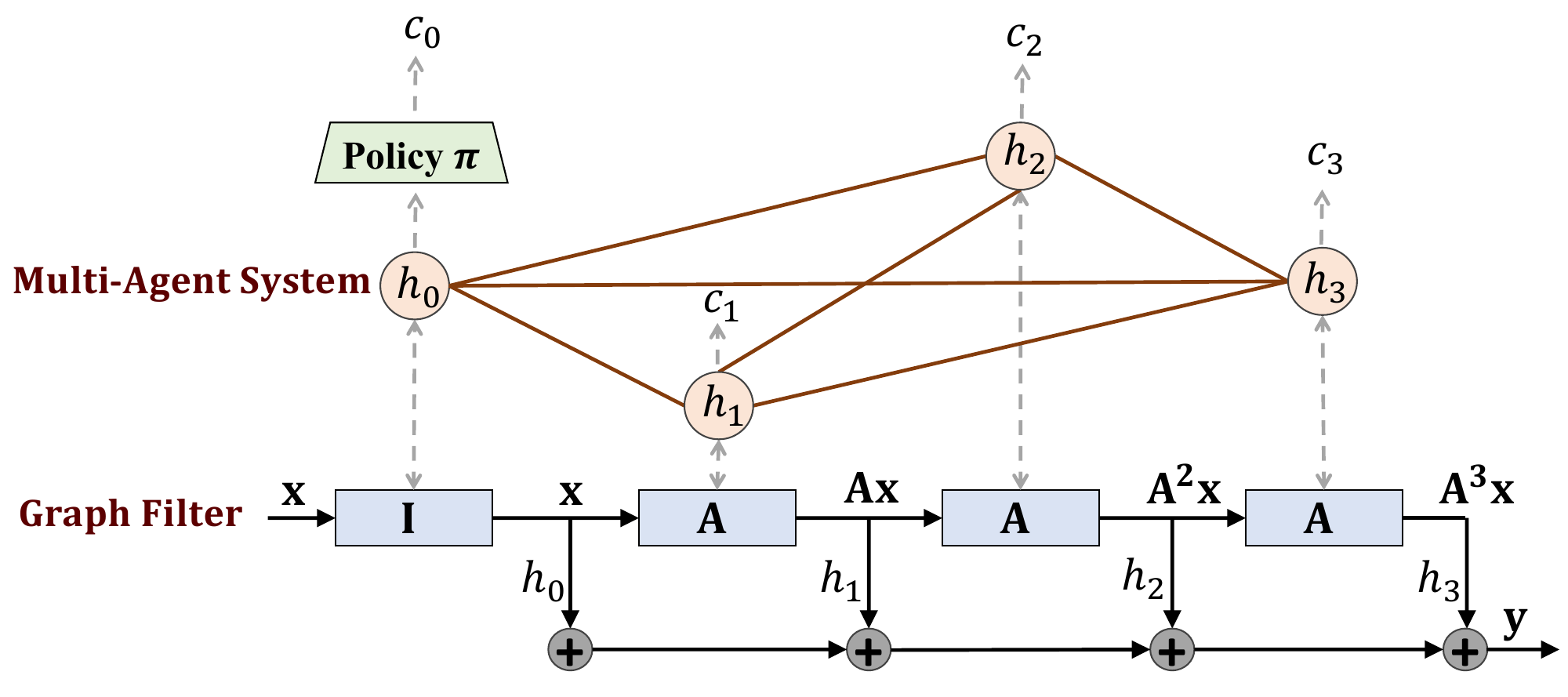}%
		\label{subfig:multAgentSystem}%
	\end{subfigure}
	\begin{subfigure}{0.85\columnwidth}
		\includegraphics[width=1.\linewidth,height = 0.425\linewidth]{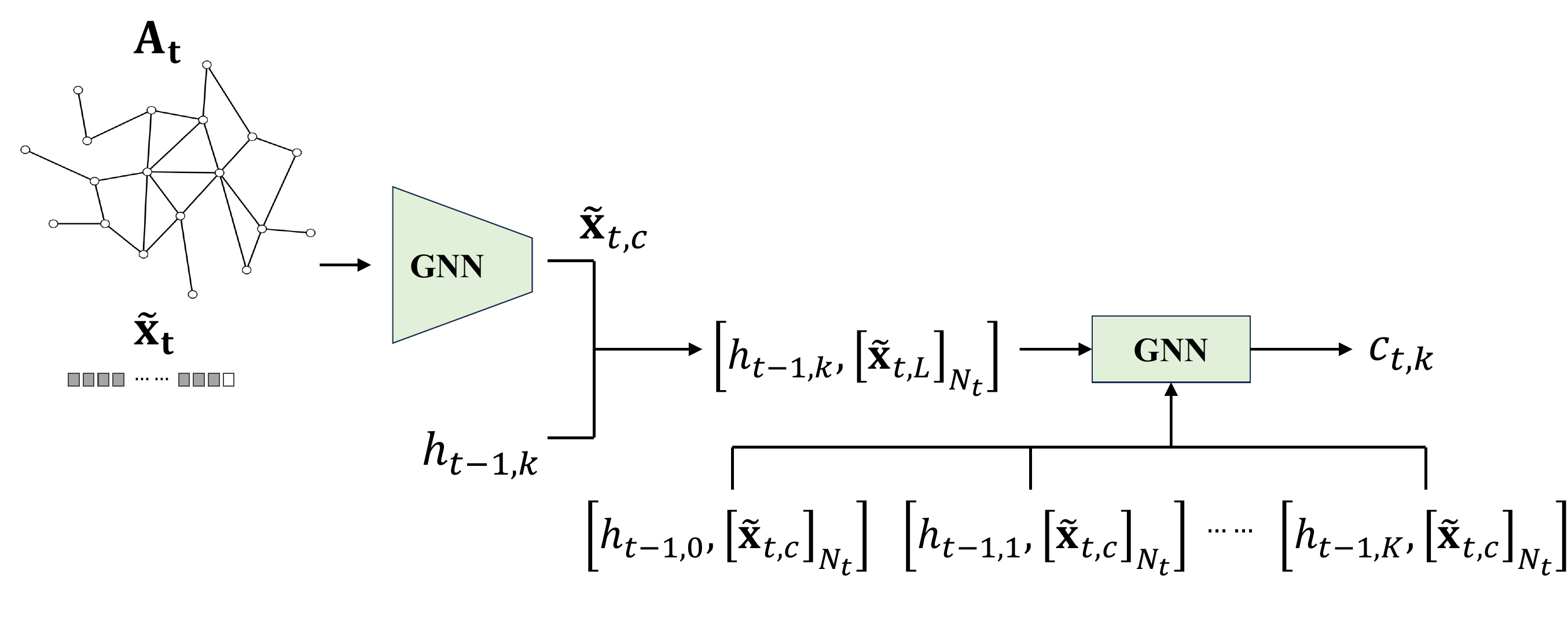}%
		\label{subfig:policy}%
	\end{subfigure}
	\caption{(Left) Framework of Graph Multi-Agent Reinforcement Learning. (Right) Context-aware Graph Neural Network.}\label{fig:framework}\vspace{-6mm}
\end{figure*} 

\subsection{Stochastic Sequential Decision Making}

Since the filter updated at a particular instant affects future inference at successive time instants, it requires accounting for long-term expansion impacts when updating the filter over time. Specifically at time $t$, consider filter parameters $\bbh_t$ as the states of the filter $\ccalF$ and the expanded graph as an external environment, where the topology $\bbA_t$ and the signal $\bbx_t$ are the environment states. We propose to design a policy $\pi$ that controls the filter across time steps, to optimize the prediction performance over incoming nodes under expansion uncertainty. This motivates to formulate a stochastic constrained optimization problem as
\begin{align}\label{eq:StoExpanding}
	\min_{\substack{\pi}} \quad &
	\mathbb{E}\Big[\sum_{t=1}^T l([\tilde{\bby}_t]_{N_t}, x_{N_{t}}) | \bbx_0, \bbA_0\Big]& \\
	\text{s.t.} \quad &
	\;\; 
	P_e(\bbA_{t}, \bbx_{t} | \bbA_{t-1}, \bbx_{t-1}, \bba_{t-1}),&\text{for}~t=1,\ldots,T \nonumber \\
	&\;\;  
	\bbh_t = \pi(\bbA_{t-1}, \bbx_{t-1}, \bba_{t-1}, \bbh_{t-1}),&\text{for}~t=1,\ldots,T
	& \nonumber \\
	&\;\;  
	\tilde{\bby}_t = \ccalF(\bbA_{t}, \tilde{\bbx}_t, \bbh_{t})~\text{[cf. }\eqref{eq:filter}\text{]},&\text{for}~t=1,\ldots,T. 
	& \nonumber
\end{align}
The policy $\pi$ determines the filter parameters $\{\bbh_t\}_{t=1}^T$ sequentially to compute the predictions $\{[\tilde{\bby}_t]_{N_t}\}_{t=1}^T$, and $P_e$ is a stochastic expansion model as described in Sec. \ref{subsec:graph}. This problem is challenging because: (C1) the expanded graph depends on the previous one stochastically, resulting in \emph{stochastic sequential decision making}; (C2) filter parameters need to be optimized based on a cumulative loss reflecting long-term impacts, instead of an instantaneous one; (C3) the graph size and problem dimension increase over time. 

We overcome these challenges by leveraging multi-agent reinforcement learning (MARL). First, we model the graph filter as a multi-agent system, where each filter shift is an agent and its parameter is the agent's state, and the expanding graph as an environment interacting with agents; together, modeling graph expansion as a Markov decision process to tackle (C1). Then, we frame \eqref{eq:StoExpanding} in the domain of MARL and learn the policy based on a cumulative long-term reward, which computes agent actions to update their states and adapts the filter to the expanding graph successively to tackle (C2). Lastly, we design a Context-aware Graph Neural Network (C-GNN) to parameterize the policy for training, which incorporates all available information to compute actions and is transferable to graphs with varying sizes to tackle (C3). 

\begin{remark}
    We consider multi-agent reinforcement learning following the nature of multi-hop filtering. Specifically, the filter shifts the signal $K$ times to collect multi-hop information and aggregates the $K$ shifted signals to extract features. Different filter orders capture individual but interconnected shifted features, and their parameters control the importance of these features. Therefore, filters correspond nicely to multi-agent systems and MARL fits well to our formulated problem.  
\end{remark}

\section{Graph Multi-Agent Reinforcement Learning}\label{sec:method}

In this section, we develop Graph Multi-Agent Reinforcement Learning (G-MARL) to solve problem \eqref{eq:StoExpanding}. 

\subsection{Multi-Agent Reinforcement Learning}\label{subsec:MARL}

For a filter of order $K$ with parameters $\{h_k\}_{k=0}^K$, we consider each filter shift $k$ as an agent $R_k$. This yields a multi-agent system $\ccalR$ of $K+1$ agents $\{R_k\}_{k=0}^K$, where the state of $R_k$ is $h_k$. By modeling the graph $\bbA$ and signal $\bbx$ as an environment $\ccalE$ that interacts with $\ccalR$, we define a Markov decision process (MDP) and re-formulate \eqref{eq:StoExpanding} in the RL domain. At time $t$, $\{R_k\}_{k=0}^K$ observe $\bbh_{t-1}$ and deploy $\pi$ to compute actions $\bbc_t$. We define $\bbc_t$ as the value changes of filter parameters, which amend $\bbh_{t-1}$ with linear dynamics as
\begin{align}\label{eq:agentDynamics}
    \bbh_{t} = \bbh_{t-1} + \bbc_t. 
\end{align}
It defines how $\pi$ controls $\bbh_t$, corresponding to the second constraint in \eqref{eq:StoExpanding}. The incoming node $v_{N_t}$ and its attachment $\bba_{t-1}$ change the environment $\ccalE_{t-1} = \{\bbA_{t-1}, \bbx_{t-1}\}$ to $\ccalE_{t} = \{\bbA_{t}, \bbx_{t}\}$ with the stochastic model in Sec. \ref{subsec:graph}. This leads to a transition probability function $P_e\big(\bbA_{t}, \bbx_{t} | \bbA_{t-1}, \bbx_{t-1}, \bba_{t-1}\big)$, depending on the stochastic dynamics of graph expansion. 

The interaction between the environment $\ccalE$ and agents $\{R_k\}_{k=0}^K$ lie in using the filter to infer the incoming signal value. We follow \eqref{eq:filter}-\eqref{eq:prediction} to predict $x_{N_t}$ as $[\tilde{\bby}_t]_{N_t}$, which feeds back a reward $l([\tilde{\bby}_{t}]_{N_{t}}, x_{N_{t}})$ that represents the instantaneous performance of agents. With a discounted factor $\gamma \in [0, 1]$ for future rewards, the expected discounted reward is 
\begin{align}\label{eq:discountedLoss}
	r(\bbx_0, \ccalG_0, \bbh_0 | \pi) = - \mathbb{E} \big[ \sum_{t=1}^{T} \gamma^{t-1} l([\tilde{\bby}_{t}]_{N_{t}}, x_{N_{t}}) \big],
\end{align}
where $\ccalG_0, \bbx_0, \bbh_0$ are initial conditions, $\mathbb{E}[\cdot]$ is w.r.t. the expansion stochasticity, and $r(\bbx_0, \ccalG_0, \bbh_0 | \pi)$ corresponds to the objective in \eqref{eq:StoExpanding}. Our goal is to find an optimal $\pi^*$ that maximizes $r(\bbx_0, \ccalG_0, \bbh_0 | \pi)$. Framing \eqref{eq:StoExpanding} from the perspective of MARL: (i) provides an explicit way how $\pi$ controls the filter adapting to the expanding graph; (ii) incorporates stochastic sequential decision making by following a MDP; and (iii) accounts for the future impact using a long-term reward \eqref{eq:discountedLoss}. 

\begin{figure*}%
	\begin{subfigure}{0.395\columnwidth}
		\includegraphics[width=1\linewidth, height = 0.8\linewidth]{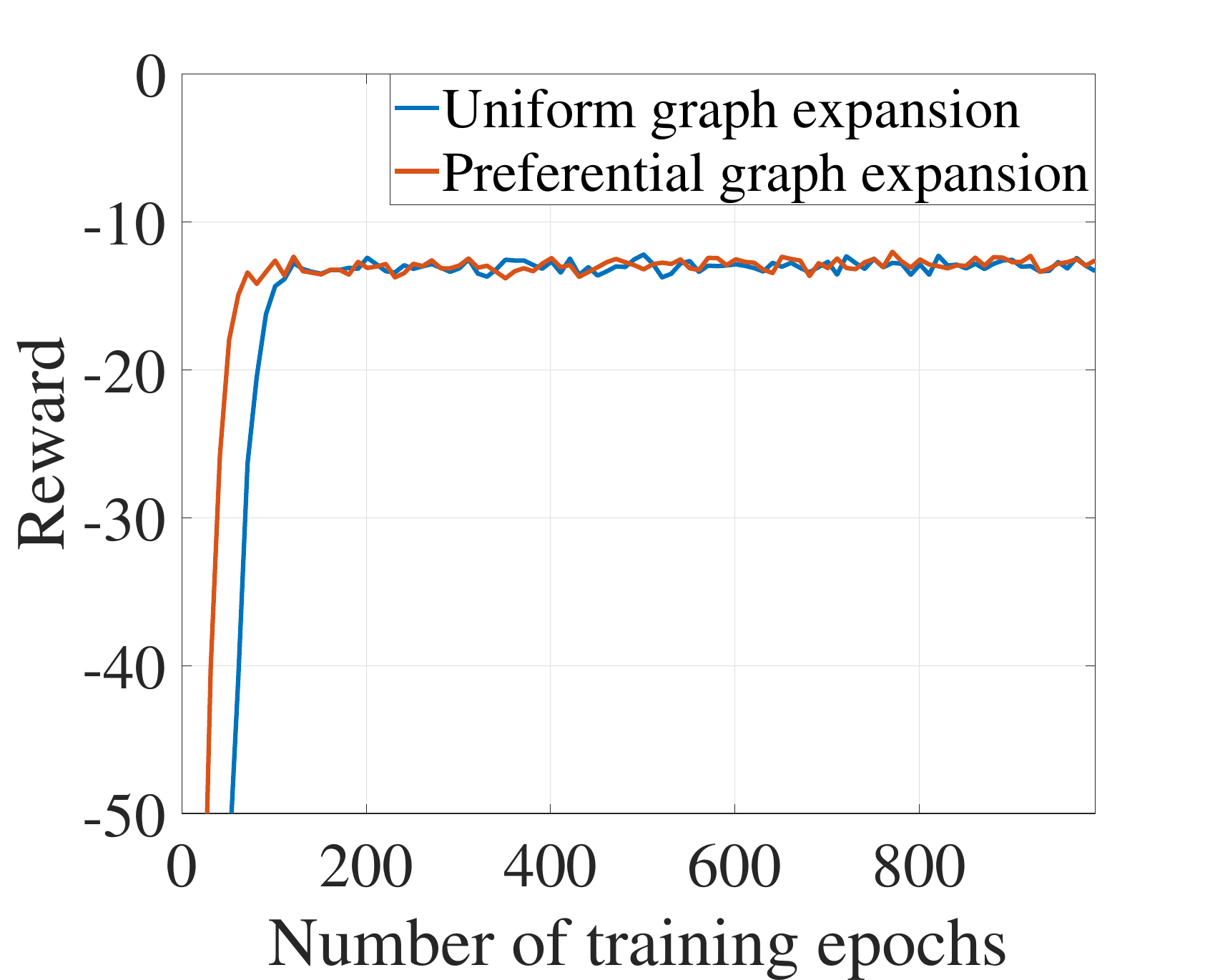}%
		\caption{}%
		\label{subfiga}%
	\end{subfigure}\hfill%
	\begin{subfigure}{0.395\columnwidth}
		\includegraphics[width=1\linewidth,height = 0.8\linewidth]{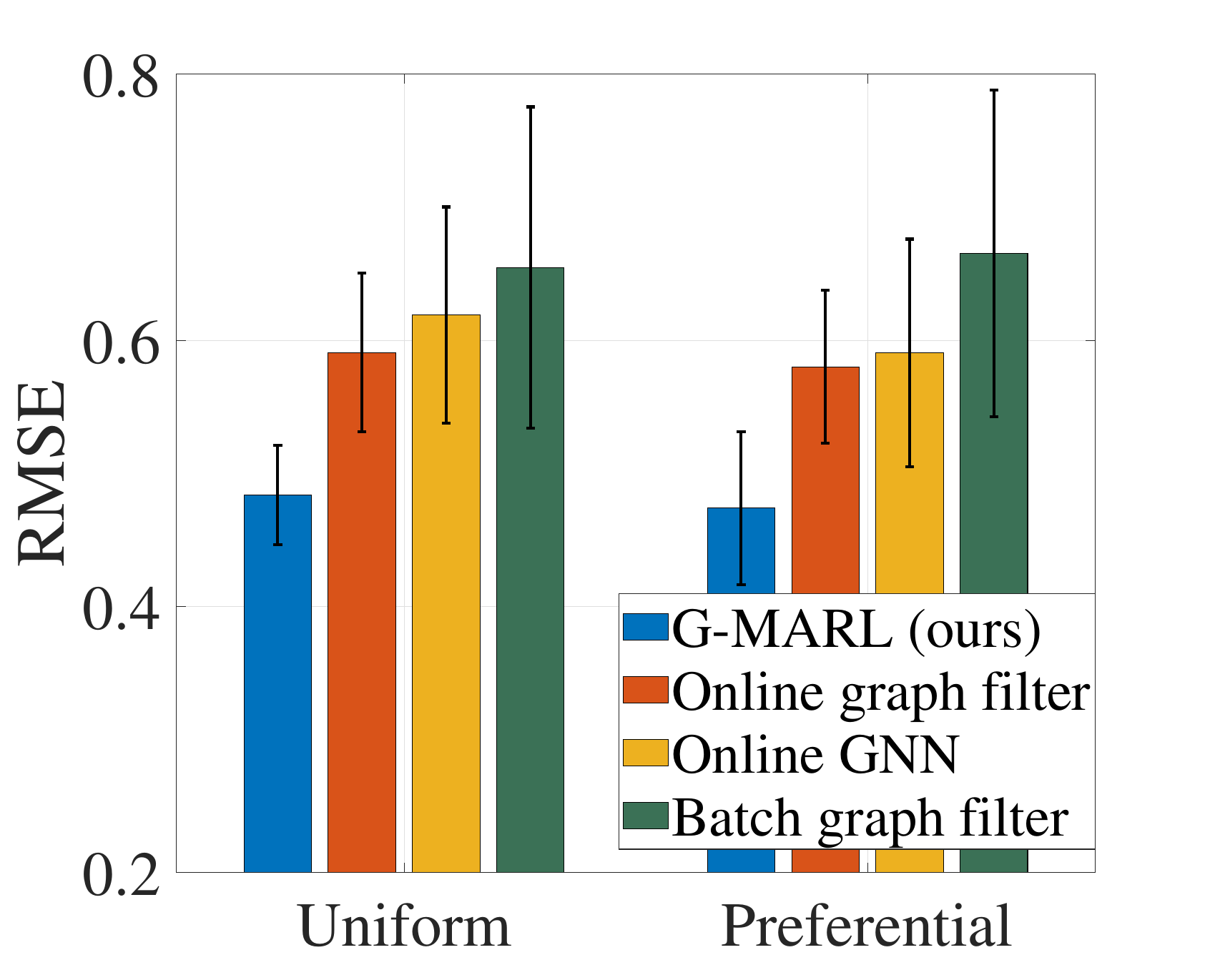}%
		\caption{}%
		\label{subfigb}%
	\end{subfigure}\hfill
	\begin{subfigure}{0.395\columnwidth}
		\includegraphics[width=1\linewidth,height = 0.8\linewidth]{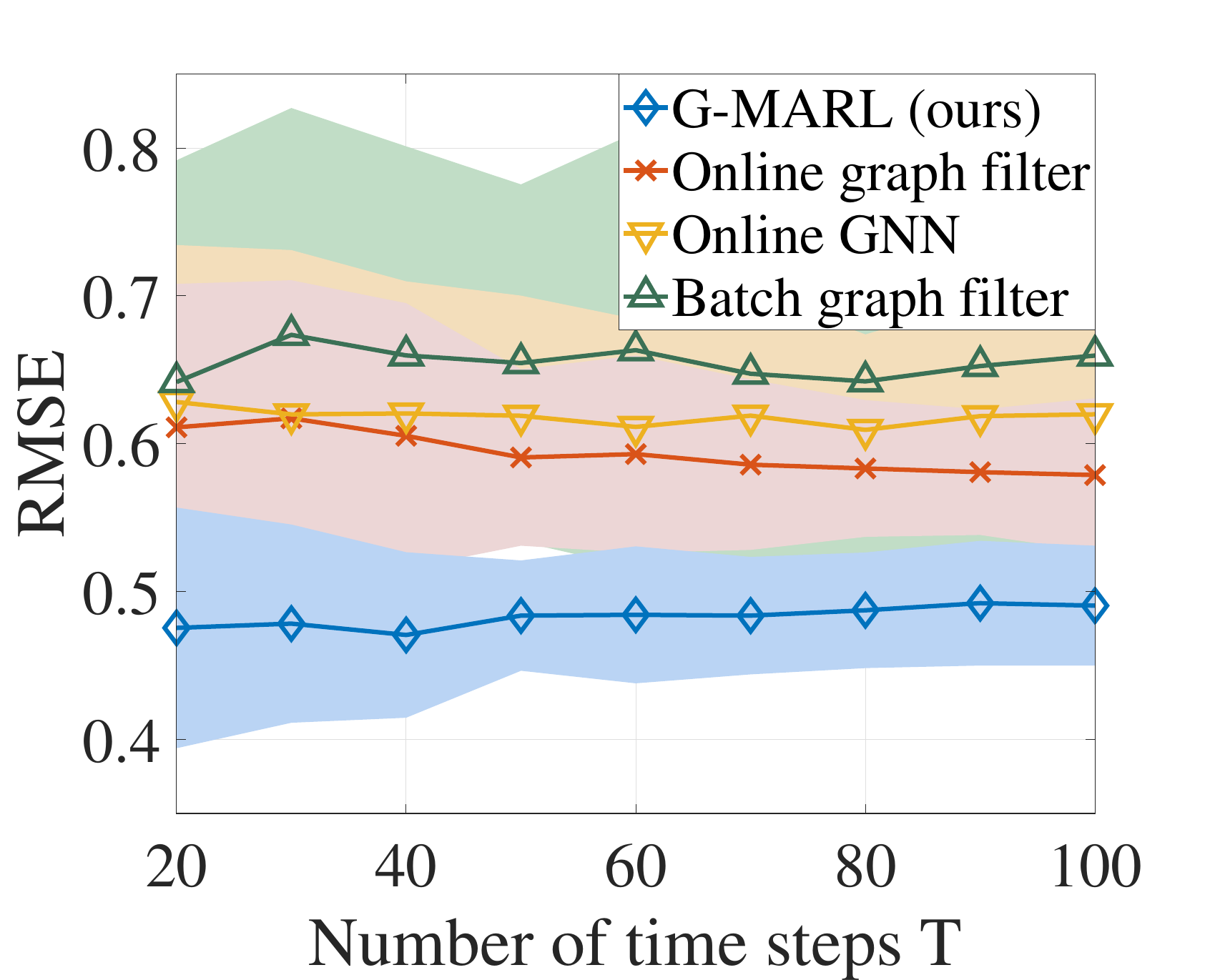}%
		\caption{}%
		\label{subfigc}%
	\end{subfigure}\hfill
	\begin{subfigure}{0.395\columnwidth}
		\includegraphics[width=1\linewidth,height = 0.8\linewidth]{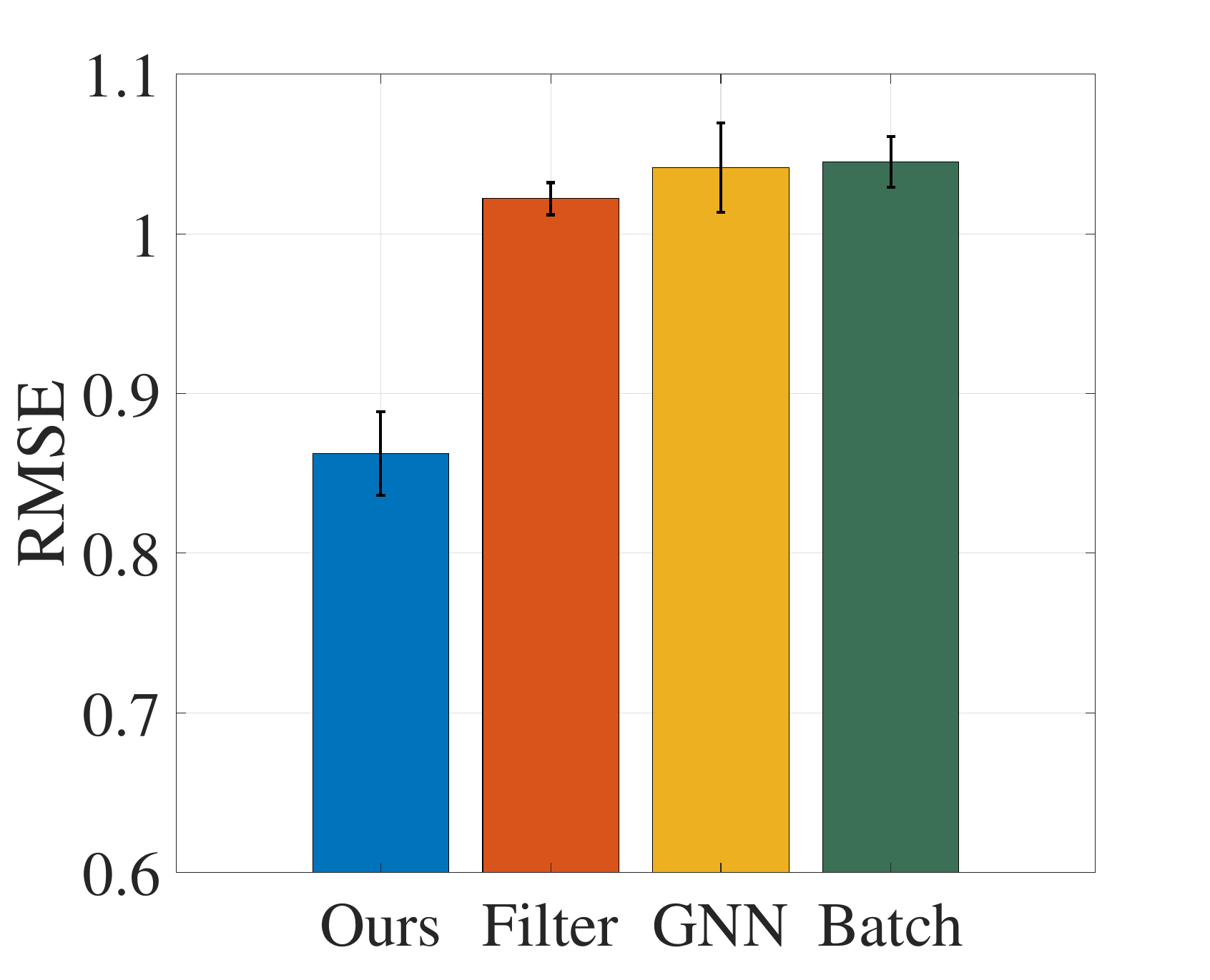}%
		\caption{}%
		\label{subfigd}%
	\end{subfigure}\hfill
	\begin{subfigure}{0.395\columnwidth}
		\includegraphics[width=1\linewidth,height = 0.8\linewidth]{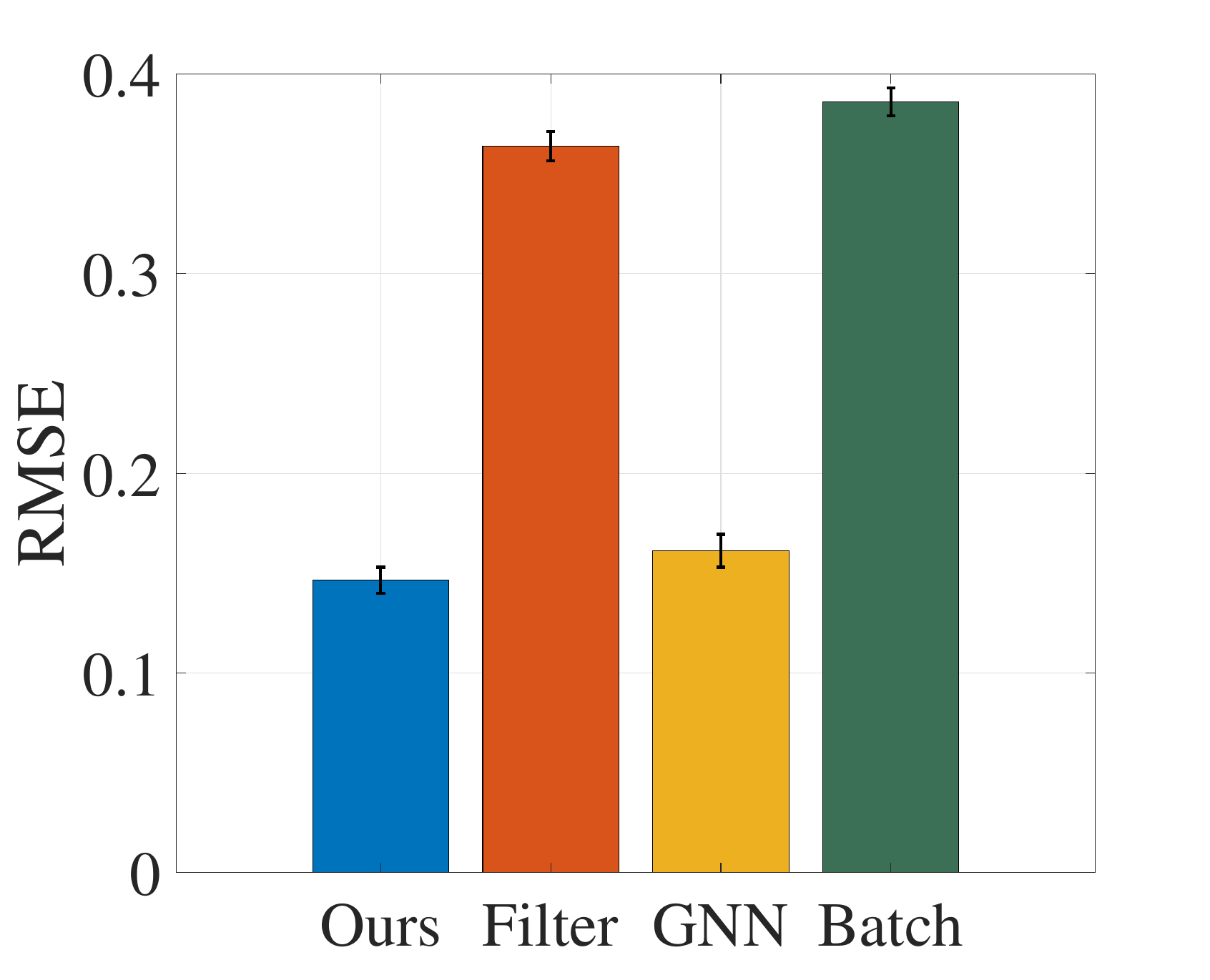}%
		\caption{}%
		\label{subfige}%
	\end{subfigure}
	\caption{(a) Training convergence of G-MARL. (b) Performance comparison for synthetic data. (c) Generalization of G-MARL to unseen scenarios. (d) Performance comparison for recommender systems. (e) Performance comparison for COVID predication.}\label{fig:performance}\vspace{-6mm}
\end{figure*}

\subsection{Context-Aware Graph Neural Networks}\label{subsec:CGNN}

We parameterize $\pi$ by a learning architecture. To leverage all information from agent states, graph topology and signal, we develop a Context-aware Graph Neural Network (C-GNN). 

At each time $t$, C-GNN \emph{first} extracts context features from the graph via the first GNN. This GNN is multi-layer with each layer comprising a filter bank and a nonlinearity, which takes the topology $\bbA_t$ and the padded signal $\tilde{\bbx}_t$ as input and outputs a context feature $\tilde{\bbx}_{t, c}$ containing information of the expanded graph \cite{gama2019convolutional}. \emph{Then}, C-GNN concatenates the context feature $[\tilde{\bbx}_{t, c}]_{N_t}$ and the agent state $h_{t,k}$ as an intermediate state $\bbz_{t,k} = [h_{t,k}, [\tilde{\bbx}_{t, c}]_{N_t}]^\top$ for agent $R_k$, which aggregates information from both the graph and the agent. \emph{Lastly}, C-GNN leverages the second GNN over the multi-agent graph to process these intermediate states, where each agent exchanges its state with other agents and aggregates these states to compute actions $\bbc_t$ \cite{gao2025co}. We represent the C-GNN as $\Phi(\bbh_{t-1}, \bbA_t, \tilde{\bbx}_t, \bbtheta) = \bbc_t$, where $\bbtheta$ are 
architecture parameters -- see Fig. \ref{fig:framework}. 

\begin{remark}
	This work focuses on filtering over expanding graphs for signal inference because it is an important task with wide applications, such as cold-start recommendation and COVID prediction, and it allows us to ease exposition in light of earlier works on graph filters. We remark that our framework is versatile that can be applied in a similar manner to other tasks in addition to inference, other processing architectures in addition to filters, and other dynamic models in addition to expansion. Our goal is to lay down the first framework to enable future research in these other settings. 
\end{remark}

\section{Numerical results}\label{sec:experiments}

We evaluate our framework in a synthetic setup based on a random expanding model and two real-data setups based on recommender systems and COVID prediction. 

\smallskip
\noindent\textbf{Synthetic setup.} Consider an initial graph with $5$ nodes and incoming nodes of length $T = 50$. The initial signal is sampled from a standard normal distribution. The graph expansion follows the Erdos–Renyi model with two heuristics \cite{mcdiarmid2020modularity}: (i) uniformly-at-random attachment, where each new node attaches to an existing node with the same probability; (ii) preferential attachment, where the probability is proportional to the degree of the existing node. The signal expansion follows a shift of the existing signal over the expanded graph with additive noise sampled from a normal distribution with variance $0.25$. The filter is of order $3$, the first GNN is a single-layer graph convolutional neural network with $32$ filters and a ReLU nonlinearity, and the second GNN is a message passing neural network with aggregation and update functions as multi-layer perceptrons of $(32, 64, 32)$ units. The learning rate is $5 \times 10^{-4}$ and the metric is root mean square error (RMSE) over time steps. Results are averaged over $64$ runs. 

Fig. \ref{subfiga} shows the training convergence of the proposed G-MARL. Both rewards increase with epochs and approach stationary solutions. Fig. \ref{subfigb} compares three baselines: (i) batch filter \cite{huang2018rating}, (ii) online filter \cite{das2024online}, and (iii) online GNN \cite{gama2020graphs}. The first establishes a quadratic program over one expansion sequence and has a closed-form least-square expression, but does not consider the expansion stochasticity and keeps filter parameters fixed throughout the expansion. The second initializes the filter with the batch solution, and updates parameters online with the instantaneous prediction loss per time step. The third uses the GNN for inference and updates parameters online in a similar manner. G-MARL outperforms significantly with the lowest RMSE because it accounts for the long-term reward over graph expansion and tunes filter parameters adapting to time-varying graphs. Online filter takes the second place as it updates parameters per time step, although considering only current (instantaneous) prediction loss. Online GNN performs slightly worse than online filter because the filter is sufficient for this problem and simpler to update online, enabling efficient adaptation to graph expansion. In contrast, GNNs are more complex and may require more extensive updates to keep pace with the expanding graph. Batch filter performs worst as it does not adapt the filter to graph changes. 

Fig. \ref{subfigc} shows the generalization capacity to unseen scenarios. We train the policy with $T = 50$, while testing it with $T \in [20, 100]$. Our method performs comparably well in unseen scenarios, despite trained in a different setting. This is because the developed C-GNN that parameterizes the policy is permutation equivariant and generalizable to unseen graphs with similar topologies, yielding stable performance. While worse than ours, online filter performs better as $T$ increases. This follows our intuition that more time steps allow more online updates, which may better adapt the filter to expanding graphs. This trend is less evident in online GNN, as its higher complexity may require more extensive updates for adaptation. 

\smallskip
\noindent\textbf{Cold-start recommendation.} This experiment generates recommendations for new users joining the system. This is a critical unsolved issue in recommender systems \cite{huang2018rating, isufi2024graph}, and we propose to tackle it by stochastic sequential decision making with graph filtering. We take the Movielens-100K dataset to perform user-based recommendation. We start with $200$ users selected at random from $943$ users, and build the initial graph $\bbA_0$ with Pearson similarity \cite{das2024online}. For each randomly incoming user, we build the attachment $\bba_t$ with $20$ edges based on its correlation w.r.t the existing users. The signal $\bbx_t$ is the collection of user ratings for the movie ``Star Wars'', which has the largest number of ratings. Fig. \ref{subfigd} shows the results. Our method achieves the best performance and its improvement is emphasized in real data, which highlight the importance of taking into account the long-term impact over graph expansion. Online filter performs better than online GNN and batch filter because it updates filter parameters tailored to the expanded graph and employs a simpler architecture with fewer parameters, making it easier for online adaptation. However, it is worse than our method because its update is only based on current (instantaneous) prediction loss. 

\smallskip
\noindent\textbf{COVID prediction.} This experiment predicts the number of COVID cases for unknown cities, and takes the data with daily cases for $269$ cities \cite{dong2020interactive}. We start with $30$ random cities and build a five nearest neighbor graph $\bbA_0$ based on pairwise distances. We also use this to compute the attachment for any incoming city, and the signal $\bbx_t$ is the collection of infection cases at day $100$. For each run, we shuffle the order in which the cities are added to the starting graph. Fig. \ref{subfige} shows that our method similarly outperforms the baselines by capturing expansion dynamics and incorporating long-term rewards. Different from previous applications, online GNN exhibits better performance than other baselines. We attribute this behavior to the stronger expressive power of GNNs, which may be needed for COVID prediction, thereby enabling faster convergence and more rapid adaptation in online learning. 

\section{Conclusions}\label{sec:conclusions}

In this work, we formulate filtering over expanding graphs as a stochastic sequential decision making problem and solve it from the perspective of multi-agent reinforcement learning. We leverage a filter to process graph signals and a policy to update filter parameters, adapting to expanding graphs. By modeling the filter as a multi-agent system, we learn the policy through MARL and develop C-GNNs for policy parameterization. The proposed framework accounts for long-term impacts over stochastic graph expansion, thereby capturing the expansion pattern and improving the performance. Experiments show its superior performance and robust generalization in both synthetic and real-world applications. This work focuses on signal-value inference using graph filters, which is a fundamental task in graph signal processing. It is worth remarking that our framework is general that can be applied to other processing models in addition to graph filters, other node-level tasks in addition to inference, and other stochastic dynamics in addition to graph expansion. Future work will investigate more complex tasks with advanced architectures. 

{\small
	\bibliographystyle{IEEEbib}
	\bibliography{strings,references}
}

\end{document}

%% file: my_sections.tex
\usepackage{needspace}

